\documentclass[10pt,twocolumn,letterpaper]{article}

\usepackage[pagenumbers]{iccv}      

\definecolor{iccvblue}{rgb}{0.21,0.49,0.74}
\usepackage[pagebackref,breaklinks,colorlinks,allcolors=iccvblue]{hyperref}
\usepackage{amssymb}
\usepackage{pifont}
\newcommand{\xmark}{\ding{55}}

\usepackage{multirow}
\usepackage{caption}

\usepackage{subcaption}
\usepackage{marvosym}

\usepackage{lmodern} 
\usepackage{xcolor} 
\usepackage{graphicx} 
\usepackage{amsmath} 
\usepackage[T1]{fontenc} 
\usepackage{inconsolata} 
\usepackage{hyperref} 

\usepackage{listings}
\usepackage{enumitem}
\usepackage[most]{tcolorbox} 

\definecolor{lightgray}{rgb}{0.95,0.95,0.95}
\definecolor{darkgray}{rgb}{0.4,0.4,0.4}
\definecolor{commentgreen}{rgb}{0.13,0.54,0.13}
\definecolor{keywordblue}{rgb}{0.18,0.18,0.57}
\definecolor{stringred}{rgb}{0.7,0.1,0.1}
\definecolor{promptbg}{rgb}{0.98, 0.98, 1.0} 
\definecolor{promptframe}{rgb}{0.8, 0.8, 0.9} 

\tcbset{
  promptblock/.style={
    enhanced,                             
    colback=promptbg,                     
    colframe=promptframe,                 
    boxrule=0.5pt,                        
    arc=3mm,                              
    outer arc=3mm,                        
    left=2mm, right=2mm, top=2mm, bottom=2mm, 
    boxsep=1mm,                           
    breakable                             
  }
}

\def\confName{ICCV}
\def\confYear{2025}

\title{HyCodePolicy: Hybrid Language Controllers for Multimodal Monitoring and Decision in Embodied Agents}

\author{
Yibin Liu$^{3,4}$\footnotemark[1] \quad
Zhixuan Liang$^{2,5}$\footnotemark[1]\hspace{2pt} \footnotemark[3]\quad
Zanxin Chen$^{5,6}$\footnotemark[1] \quad
Tianxing Chen$^{2,6}$ \quad
Mengkang Hu$^{2}$\\
Wanxi Dong$^{7}$ \quad
Congsheng Xu$^{1}$ \quad
Zhaoming Han$^{4}$ \quad
Yusen Qin$^{4,8}$ \quad
Yao Mu$^{1,5}$\footnotemark[2]\\
$^1$SJTU ScaleLab \quad
$^2$HKU MMLab \quad
$^3$NEU \quad
$^4$D-Robotics \quad
$^5$Shanghai AI Lab\\
$^6$SZU \quad
$^7$SUSTech \quad
$^8$THU \\
{\small
\footnotemark[1]\hspace{2pt} Equal Contribution \quad
\footnotemark[3]\hspace{2pt} Project Lead \quad
\footnotemark[2]\hspace{2pt} Corresponding Author}\\
{\tt\small liuyibin@stumail.neu.edu.cn, zxliang@cs.hku.hk, yaomarkmu@gmail.com}
}

\begin{document}
\maketitle

\begin{abstract}
Recent advances in multi-modal large language models (MLLMs) offer powerful perceptual grounding for code policy generation in embodied agents. However, most existing systems lack effective mechanisms to adaptively monitor execution and iteratively repair policies in response to failures. In this work, we introduce \textit{HyCodePolicy}, a hybrid language-based control framework that closes the loop between code synthesis, geometry-aware grounding, perceptual monitoring, and targeted repair. Given a natural language instruction, our system first decomposes it into hierarchical sub-goals and generates an initial program grounded in object-centric geometric primitives. HyCodePolicy then executes the program in simulation, with a vision-language model (VLM) monitoring designated checkpoints to identify and localize failures and inferring their underlying causes. By integrating structured execution logs that capture program-level events with VLM-derived perceptual feedback, HyCodePolicy pinpoints root causes of failures and applies targeted code repairs. This hybrid dual feedback mechanism enables self-correcting program synthesis with minimal human intervention. Our results demonstrate that HyCodePolicy significantly enhances the robustness and sample efficiency of robot manipulation policies, offering a scalable strategy for incorporating multi-modal reasoning into autonomous decision-making pipelines.
\end{abstract}
    
\section{Introduction}
\label{sec:intro}

\begin{figure}
    \centering
    \includegraphics[width=1\linewidth]{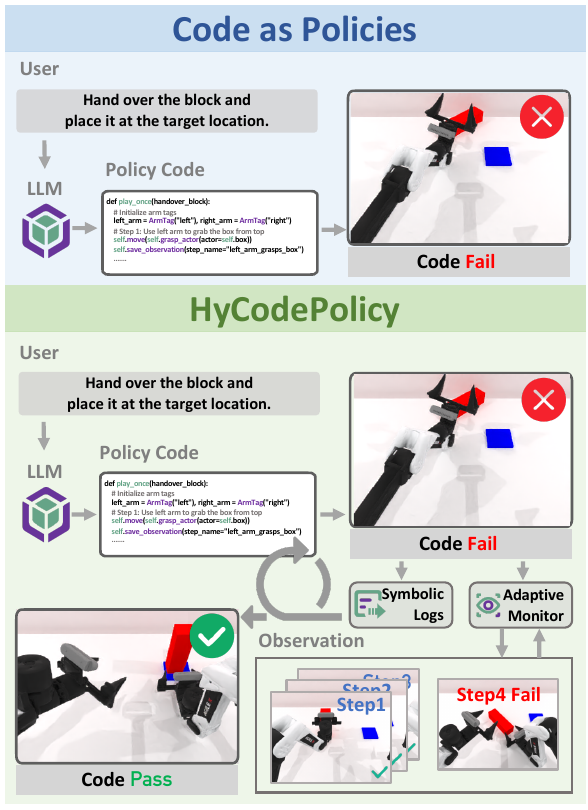}
    \caption{
Overview figure of \textbf{HyCodePolicy}, a closed-loop framework for language-conditioned manipulation with hybrid program synthesis, monitoring, and repair.
}
    \label{fig:enter-label}
    \vspace{-8pt}
\end{figure}

The burgeoning capabilities of large language models (LLMs) and their multi-modal counterparts (MLLMs) are rapidly transforming the landscape of artificial intelligence, opening unprecedented avenues for robot planning. From this broad perspective, we focus on language-grounded manipulation, where robots leverage LLMs to interpret high-level natural language instructions, reason about complex tasks, and execute them in physical environments~\cite{saycan2022arxiv, jiang2022vima}. This paradigm provides an opportunity to democratize robot programming, moving beyond tedious explicit coding to intuitive linguistic instruction. However, a fundamental challenge persists that effectively bridging the rich semantic expressiveness of natural language with precise, structured, and physically grounded representations for reliable robot execution.

Prior works have made considerable strides in synthesizing robot actions from language, ranging from direct LLM-generated plans to formal logic representations (\textit{e.g.} code-as-policy) approaches~\cite{codeaspolicies2022, singh2023progprompt, liu2023llm+p}. These methods typically commit to a one-shot generation of the complete behavior plan, relying entirely on that single attempt being correct. Yet, real-world robotic tasks are inherently uncertain due to perception noise, execution errors, and dynamic environments. Current systems frequently lack robust mechanisms to adaptively monitor task execution, detect and diagnose failures, and then repair the robot's behavior in a closed loop~\cite{jiang2025feedback}. This limitation undermines the robustness, efficiency, and real-world reliability of language-conditioned robot policies, often requiring extensive human intervention for debugging and recovery.

To address this critical gap, we introduce \textbf{HyCodePolicy}, a novel hybrid language-based control framework designed for robust, self-correcting robotic manipulation. HyCodePolicy systematically unifies code synthesis, geometric grounding, multi-modal monitoring within a closed-loop programming cycle. Rather than treating generated code as a static output, it treats each program as an evolving hypothesis that can be actively validated, evaluated, and corrected via perceptual cues and symbolic reasoning.

HyCodePolicy comprises four synergistic components: (1) High-level language intent grounding through hierarchical subgoal decomposition and geometrically informed program synthesis; (2) Simulated execution coupled with symbolic logging and concurrent vision-language model (VLM) observations; (3) Hybrid failure attribution fusing symbolic and perceptual diagnostics to infer causal error hypotheses; and (4) Iterative program repair, achieving the closed-loop control via targeted and interpretable code updates. This hybrid feedback mechanism enables self-correcting program synthesis with minimal human supervision.

We conduct extensive experiments on RoboTwin Platform~\cite{mu2024robotwin} demonstrating the effectiveness of HyCodePolicy and showing the improvement of task success rates from 47.4\% to 63.9\% and from 62.1\% to 71.3\% in different settings. Additionally, HyCodePolicy reduces convergence iterations from 2.42 to 1.76, highlighting its ability to improve both robustness and efficiency in dynamic, real-world environments. These results underscore the practical impact of our framework in enhancing robotic manipulation tasks.  

Our key contributions are:
\begin{itemize}
    \item \textbf{A Novel Closed-Loop Control Framework:} We propose \textbf{HyCodePolicy}, a pioneering architecture that seamlessly integrates language-conditioned program synthesis with adaptive multimodal monitoring and iterative repair, improving the robustness and self-correction of robot policies.
    \item \textbf{Hybrid Grounded Feedback for Causal Repair:} We develop a unique hybrid feedback mechanism that fuses symbolic execution logs with Vision-Language Model (VLM)-based perceptual observations. This enables precise, causally-grounded failure attribution and drives targeted code repair, supported by geometric primitives for physically executable policies.
    \item \textbf{Demonstrated Robustness and Efficient Interface:} We empirically demonstrate that HyCodePolicy significantly enhances the robustness and sample efficiency of robot manipulation policies across diverse tasks. This is further demonstrated in \textbf{Bi2Code}, a re-engineered modular interface designed to optimize structured prompting and multimodal tracing for effective deployment.
\end{itemize}
The remainder of this paper is organized as follows: Section~\ref{section:related_work} reviews related work on language grounding for robotics and LLM-guided program repair. Section~\ref{section3.2expert-data-gen} details the design and implementation of HyCodePolicy. Section~\ref{section:experiment} presents our experimental setup and empirical results. Finally, Section~\ref{section:conclusions_limitations} concludes and discusses future directions. Our code is open-sourced at \href{https://github.com/RoboTwin-Platform/RoboTwin/tree/main}{RoboTwin-Platform}.
\section{Related Work}\label{section:related_work}

\subsection{Robotic Manipulation Planning with Language Grounding}

The integration of large language models (LLMs) into robotic manipulation has led to significant advancements in language-conditioned planning~\cite{saycan2022arxiv,huang2024rekep,mu2023embodiedgpt,skilldiffuser,mu2024robocodex,dexhanddiff,ni2023metadiffuser,hu2024hiagent,hu2025text2world,chen2025benchmarking}. Bridging the gap between the semantic expressiveness of language and the structured representations remains a critical challenge for LLM-based task planning.

Recent research focuses on symbolic and embedding-based approaches. Firstly, symbolic methods, such as those discussed by Cohen et al.~\cite{cohen2024survey}, provide interpretability and enforce constraints, while embedding-based approaches offer generality but lack transparency. The Embodied Agent Interface~\cite{li2024embodied} introduces a standardized framework for integrating LLMs with robotic agents. Text-to-plan techniques like SayCan~\cite{saycan2022arxiv} and Lang2LTL~\cite{liu2022lang2ltl} focus on converting language into structured plans, while Code-as-Symbolic-Planner~\cite{chen2025code} and GenCHiP~\cite{chen2025code} focus on constraint-compliant policy generation.

Moreover, language-to-program pipelines, such as Code-as-Policies~\cite{codeaspolicies2022} and ProgPrompt~\cite{singh2023progprompt}, aim to enhance robot code generation for grounded executability. Additionally, multi-modal methods, like VIMA~\cite{jiang2022vima} and EmbodiedGPT~\cite{mu2023embodiedgpt,mu2024robocodex,chen2024roboscript}, combine vision and language to enable generalization across diverse tasks.

HyCodePolicy extends these approaches by integrating structured program synthesis with symbolic-perceptual feedback for closed-loop planning and self-correction, as described in Section~\ref{section3.2expert-data-gen}.

\subsection{MLLM-Guided Failure Diagnosis and Program Repair}

Studies on MLLM-guided failure attribution focus on feedback-driven model refinement. Approaches like Self-Debugging~\cite{chen2023teaching} and Self-Refine~\cite{madaan2023self,reinforced-active} explore iterative correction through self-generated explanations, but they are tested often only under idealized conditions~\cite{jiang2025feedback}.

Executable program interfaces have been developed for failure attribution and repair. CodeAct~\cite{wang2024executable} and INTERVENOR~\cite{wang2024intervenor} enhance multi-turn repair by modeling agent reasoning as code. Safety-focused methods, such as SafetyChip~\cite{yang2024plug} and SAFER~\cite{khan2025safety}, incorporate formal reasoning to enforce task constraints and ensure safety.

HyCodePolicy enhances failure localization and repair by fusing multimodal perceptual feedback with symbolic state, supporting interpretable, robust and adaptive behavior in real-world tasks. Unlike prior work relying on introspection or static logic, our method integrates dynamic feedback to drive targeted repair.
\section{Method}\label{section3.2expert-data-gen}

In contrast to prior Code-as-policy frameworks that treat program generation as a one-shot synthesis process, HyCodePolicy introduces a flexible, closed-loop architecture in which code becomes not only an execution medium but also a vehicle for perception, self-monitoring, and autonomous refinement. The core insight is to reinterpret the generated program as an evolving hypothesis—one that is subject to empirical validation and continuous revision.

\begin{figure*}[h]
\centering
\includegraphics[width=0.98\linewidth]{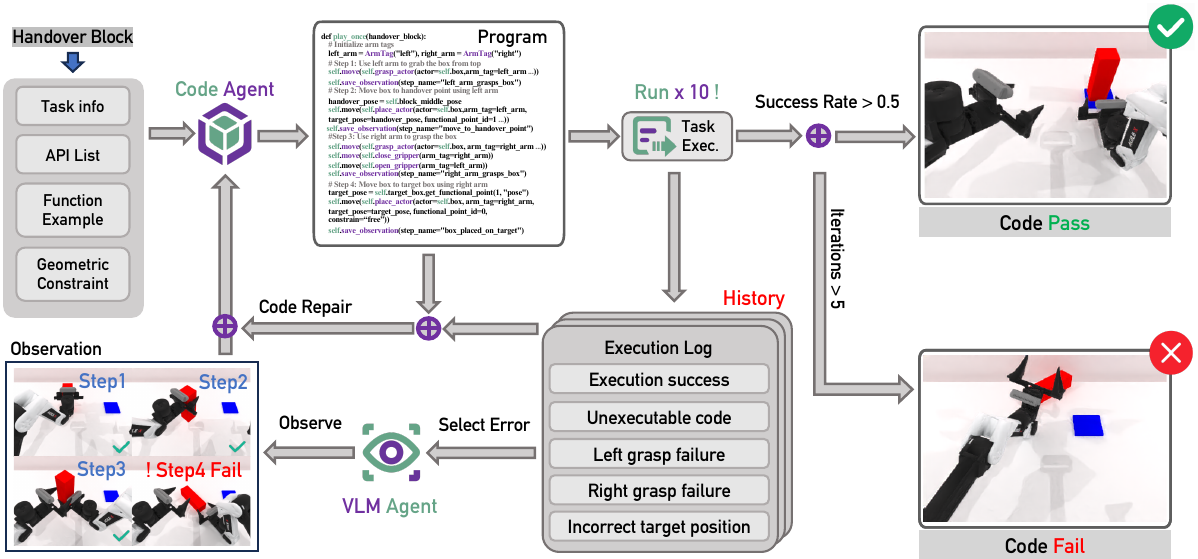}
\caption{\textbf{HyCodePolicy: Expert Code Generation Pipeline.} The pipeline integrates language-conditioned program synthesis with multimodal monitoring and iterative repair, enabling adaptive and self-correcting robotic behaviors. It combines high-level task grounding, simulated execution with feedback-driven diagnostics, and a closed-loop repair cycle to refine robot policies over time.}
\vspace{-2pt}
\label{fig:gpt_gen}
\end{figure*}

This perspective enables two key properties: \textit{falsifiability} and \textit{evolvability}. A program is falsifiable in the sense that its execution within a simulated environment exposes its limitations, such as geometric infeasibility or logical contradictions, which can be detected through runtime perceptual signals. It is evolvable because it participates in a monitor-diagnose-repair cycle, allowing it to self-correct and improve over time. This reframing supports a feedback-driven programming loop, where code actively engages in its own refinement, effectively realizing \textit{code-as-monitor} within the broader system.

As shown in Figure~\ref{fig:gpt_gen}, HyCodePolicy comprises four tightly integrated phases: (1) \textbf{Grounding high-level language intent} through hierarchical subgoal decomposition and geometrically informed program synthesis; (2) \textbf{Simulated execution and symbolic-perceptual monitoring} with structured logging and vision-language observation; (3) \textbf{Hybrid failure attribution} through fusion of symbolic and visual diagnostics; and (4) \textbf{Closed-Loop Autonomy via Adaptive Monitoring and Iterative Code Evolution} that achieves the closed-loop control via targeted and interpretable code updates.

\subsection{Grounding High-Level Intent in Code}

\textbf{HyCodePolicy} grounds high-level task intent by decomposing natural language instructions into structured subgoals and synthesizing executable Python programs aligned with geometric affordances. Given a language instruction (e.g., \textit{``Handover the block''}), a task name, and optional examples, a language model $\mathcal{L}$ produces a sequence of $N$ semantically coherent subgoals:
\[
\mathcal{S} = \{s_1, s_2, \ldots, s_N\} = \mathcal{L}(T)
\]
Each subgoal $s_i$ represents a high-level behavioral unit (e.g., ``pick up the blue block'') and serves as a constraint for subsequent program synthesis.

Codes are generated through structured prompting conditioned on three elements: a general API list, exemplar function calls, and subgoal constraints. This frames the synthesis process as a constrained, structured prediction task over the program space, enabling functional and syntactic validity.

To ensure the resulting programs are not only logically coherent but also physically executable, the system integrates geometric reasoning via a library of \textbf{Geometric Operation Primitives}. These primitives abstract physical constraints into two representational categories:

\vspace{4pt}
\noindent \textbf{Point Operation Primitives} ($\mathcal{P}$): define key spatial targets:
\begin{itemize}[leftmargin=12pt, topsep=1pt]
    \item[] $p_{\text{grasp}}$: Stable grasp point.
    \item[] $p_{\text{place}}$: Placement support point.
    \item[] $p_{\text{util}}$: Interaction site for functional use.
\end{itemize}

\vspace{4pt}
\noindent \textbf{Axis Operation Primitives} ($\mathcal{A}$): capture directional alignment:
\begin{itemize}[leftmargin=12pt, topsep=1pt]
    \item[] $a_{\text{grasp}}$: Approach axis for grasping.
    \item[] $a_{\text{place}}$: Orientation axis for placing.
    \item[] $a_{\text{util}}$: Motion axis for functional tasks.
\end{itemize}

\vspace{4pt}
These abstractions guide the synthesis of subgoal-conditioned code that respects geometric feasibility. For example, relative pose constraints ensure robust grasps and precise placements. By embedding geometric priors into code synthesis, the system reduces execution-time failures and grounds language instructions in the spatial structure of the environment.

\subsection{Simulate Execution \& Multimodal Monitoring}

\subsubsection{Program Execution and Symbolic Logging}

Once an initial program is synthesized, it is executed within a simulated robotic environment to validate its operational correctness. Each program undergoes ten independent trials to account for stochasticity in robot control, physics simulation, and sensory input. After each batch of executions, the system produces structured symbolic logs that record the outcome (success/failure) of each run, along with diagnostic error messages categorizing failure types---such as unreachable grasp configurations, invalid function calls, or incorrect placements.

These symbolic logs serve as a low-level feedback channel, capturing the syntactic and functional integrity of the program. However, they are inherently limited in attributing failures that arise from subtle visual or semantic inconsistencies.

\subsubsection{Concurrent Multimodal Observation}

To supplement symbolic logs with richer perceptual insight, we introduce a vision-language model (VLM) agent that monitors execution in parallel. This component plays a dual role that it observes and records critical state transitions, and analyzes these transitions to assess sub-goal completion.

Observation points are strategically inserted by analyzing the program structure for each subgoal $s_i \in \mathcal{S}$. For each subgoal, we identify a set of operations $\mathcal{O}^i = \{o^i_1, \ldots, o^i_{K_i}\}$, and apply a filtering function $\phi$ defined as:
\[
\phi(o^i_k) = 
\begin{cases}
1 & \text{if } o^i_k \text{ causes a visible state change} \\
0 & \text{otherwise}
\end{cases}
\]

Whenever $\phi(o^i_k) = 1$, an observation function \texttt{save\_camera\_images()} is invoked to capture the visual context post-operation. Observations are also collected at the beginning ($t = 0$) and end ($t = T$) of execution. The complete set of visual observations is:
\[
\mathcal{V} = \{v_0\} \cup \{v^i_k \mid \phi(o^i_k) = 1\} \cup \{v_T\}
\]

Each $v$ includes RGB-D images, timestamps, step identifiers, and the associated program context, enabling fine-grained alignment between visual evidence and symbolic execution steps.

\subsection{Hybrid Feedback and Failure Attribution}

\subsubsection{VLM-based Perceptual Verification}

Following program execution, the VLM agent analyzes the sequence of collected observations to determine whether each subgoal was successfully completed. For each $s_i$, the model evaluates the corresponding visual frames $v^i_{1:K_i}$ and returns a binary success signal:
\[
\hat{y}_i = \text{Observation}(v^i_{1:K_i}) \in \{0, 1\}
\]

If $\hat{y}_i = 1$, the subgoal is considered complete. Otherwise, the model initiates a failure analysis routine.

In the case of a failure, the VLM identifies the precise point of deviation $t^*_i$ within the subgoal’s execution and infers a high-level causal hypothesis $c_i \in \{\text{logic error}, \text{API misuse}, \text{execution failure}, \ldots\}$. This diagnosis provides a semantically meaningful interpretation of the error, rooted in perceptual context.

\subsubsection{Fusing Symbolic and Perceptual Feedback for Diagnosis}

By fusing the VLM’s perceptual diagnostics with the symbolic logs obtained during simulation, the system produces a joint interpretation of the failure. Symbolic traces provide procedural integrity checks, while visual diagnostics localize failures in space and time and characterize their nature (e.g., incorrect grasp angle, missing object alignment).

This hybrid diagnosis is critical for transitioning from mere detection to causal understanding. The fused feedback is encoded as a structured signal that conditions the next stage of program revision. It enables the system to isolate problematic operations and prioritize repairs according to semantic relevance and execution risk.

\subsection{Closed-Loop Autonomy via Adaptive Monitoring and Iterative Code Evolution}

\subsubsection{Adaptive Monitoring via Selective Observation and Log-Guided Re-inspection}

A defining feature of \textbf{HyCodePolicy} is its ability to make adaptive decisions on when and where to deploy multimodal perception. This adaptivity unfolds along two axes: \textit{program-level observation insertion} and \textit{execution-level trial selection}.

First, during initial synthesis, HyCodePolicy selectively inserts observation hooks based on the structure and semantics of the generated code. Specifically, operations that are likely to induce visually observable changes—such as object displacement, alignment-sensitive placements, or grasp transitions—are tagged for post-execution image capture. This avoids unnecessary monitoring overhead while ensuring coverage of visually informative transitions.

Second, across ten stochastic executions of the same candidate program, HyCodePolicy aggregates symbolic logs to identify the trial with the most diagnostically salient failure. Rather than uniformly analyzing all executions, the system selects a single representative trial for multimodal diagnosis:
\[
i^* = \arg\max_i \, \psi(\text{FailureSeverity}_i, \text{TraceDivergence}_i)
\]
where $\psi$ is a scoring function that prioritizes executions exhibiting severe failure modes and divergent symbolic traces. Visual inspection is then triggered exclusively for this most informative instance, ensuring efficient use of perceptual resources.

 \begin{figure}[t]
     \centering
     \includegraphics[width=1\linewidth]{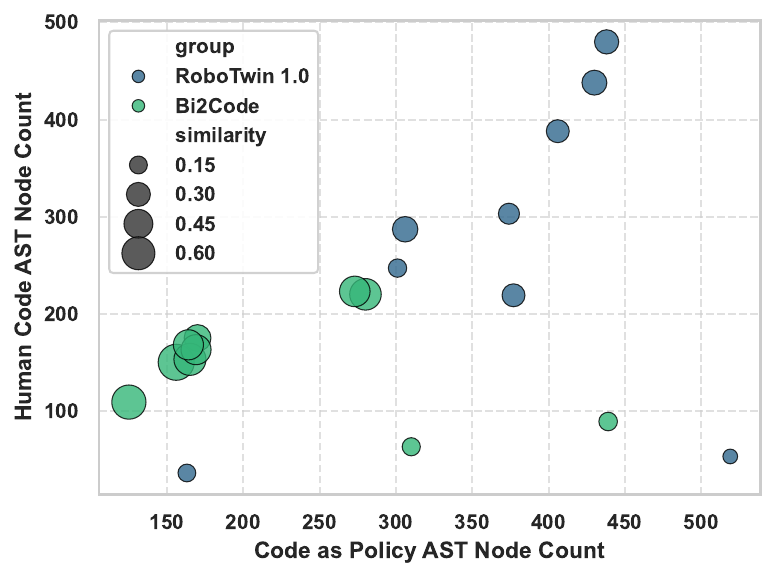}
     \vspace{-20pt}
     \caption{\textbf{Distribution of AST similarity and node counts comparing robotic manipulation code generated by RoboTwin 1.0 and Bi2Code with human-written code.} Dot size indicates structural similarity; color denotes source group.}
     \label{fig:enter-label}
     \vspace{-8pt}
 \end{figure}

Together, these mechanisms constitute an adaptive attention mechanism over both code structure and execution history, allocating diagnostic effort to the spatiotemporal loci of maximum uncertainty.

\begin{table*}[t]
\small
  \centering
  \vspace{2pt}
  \begin{tabular}{lccc}
    \toprule
    \textbf{Cond.} & \textbf{ASR} & \textbf{Top5-ASR} & \textbf{CR-Iter} \\
    \midrule
    Code as Policies w. RoboTwin 1.0     & 47.4\% & 57.6\% & 1.00  \\
    CodeAct w. RoboTwin 1.0              & 60.4\% & 71.4\% & 2.46  \\
    HyCodePolicy w. RoboTwin 1.0         & 63.9\% & \textbf{74.2\%} & 2.42 \\
    \midrule
    Code as Policies w. Bi2Code          & 62.1\% & 68.0\% & 1.00  \\
    CodeAct w. Bi2Code                   & 66.7\% & 73.6\% & 1.89 \\
    HyCodePolicy w. Bi2Code              & \textbf{71.3\%} & \textbf{78.6\%} & \textbf{1.76} \\
    \bottomrule
  \end{tabular}
  \vspace{-3pt}
\caption{\textbf{Overall Performance Comparison of Language-Conditioned Robotic Policy Generation Frameworks.} Average Success Rate (ASR), Top-5 ASR, and Mean Code Revision Iterations (CR-Iter) are reported for different policy generation and repair frameworks (Code as Policies, CodeAct, HyCodePolicy) across RoboTwin 1.0 and Bi2Code interfaces.}
  \label{tab:robotwin_results}
  \vspace{-3pt}
\end{table*}

\begin{table}[t]
  \centering
  \small
  \begin{tabular}{lcc}
    \toprule
    \textbf{Metric} & \textbf{RoboTwin 1.0} & \textbf{Bi2Code} \\
    \midrule
    Prompt Token Length ↓       & 5901.0 & \textbf{4719.1} \\
    Code Token Length ↓         & 1236.6 & \textbf{569.4} \\
    Parallelism Control ↑               & \xmark & \checkmark \\
    AST Similarity~\cite{wen2019code} ↑ & 23.72\% & \textbf{44.78\%} \\
    CodeBLEU Similarity~\cite{ren2020codebleu} ↑            & 17.18\% & \textbf{18.53}\% \\
    CodeBERT Similarity~\cite{feng2020codebert} ↑        & 97.72\% & \textbf{98.80\%} \\
    Unixcoder Similarity~\cite{guo2022unixcoder} ↑        & 76.24\% & \textbf{82.21\%} \\
    \bottomrule
  \end{tabular}
  \vspace{-3pt}
  \caption{\textbf{Code Generation Efficiency and Quality Comparison.} Evaluation of prompt and generated code characteristics, along with code similarity metrics (AST Structural Similarity, CodeBERT, and Unixcoder cosine similarity) against expert-written human code, for RoboTwin 1.0 and Bi2Code in zero-shot generation.}
\label{tab:codegen_efficiency_transposed}
\vspace{-5pt}
\end{table}

\subsubsection{Closed-Loop Repair and Policy Evolution}

Upon fusing symbolic and visual diagnostics, HyCodePolicy initiates a targeted repair cycle. Faulty operations are localized, and the code-generation agent proposes structured edits based on the failure mode, ranging from logic rewrites and API substitutions to geometric parameter retuning. Repairs are constrained by a symbolic grammar and subgoal template, preserving compatibility and ensuring downstream executability.

This diagnosis-driven correction process is iterated in a closed loop. Each revised program is re-executed, re-monitored, and re-diagnosed, forming a feedback-driven refinement pipeline. Over multiple iterations, policies evolve into stable, interpretable, and perceptually grounded solutions. Crucially, this evolution is not static fine-tuning but an active restructuring of the program in response to empirical failures.

The result is a form of \textit{policy evolvability}: the ability of generated programs to improve autonomously through multimodal self-assessment and revision. Unlike static one-shot approaches, HyCodePolicy yields adaptive controllers that grow more robust through iterative experience, embodying a scalable strategy for long-horizon task acquisition under uncertainty.

\section{Experiment}\label{section:experiment}

\subsection{Experimental Setup}

We evaluate our framework on a shared suite of 10 robotic manipulation tasks supported by both RoboTwin~1.0~\cite{mu2024robotwin,Mu_2025_CVPR} and our redesigned Bi2Code interface, which is built upon RoboTwin2.0\cite{chen2025robotwin}. Each task is defined via a natural language instruction and executed in a physics-based simulation environment. For each configuration, the code-generation agent synthesizes 10 candidate programs per task, each executed 10 times. Results are averaged to mitigate stochasticity in perception and physics.

Our approach integrates \textit{DeepSeek-V3} for program synthesis and \textit{moonshot-v1-32k-vision-preview} for multimodal observation and diagnosis. We consider three hierarchical configurations. The detailed prompt structures, code templates, and environment metadata used in each configuration are provided in Appendix~\ref{sec:appendix_prompt}.

\begin{itemize}
    \item \textbf{Code as Policies:} One-shot generation with no feedback. This baseline reflects a static mapping from instruction to program.
    \item \textbf{CodeAct:} Symbolic feedback and trace-driven repair.
    \item \textbf{HyCodePolicy:} Our full closed-loop pipeline that integrates both symbolic and vision-language feedback for perceptually grounded repair.
\end{itemize}

To enable effective deployment of HyCodePolicy, we reengineered RoboTwin1.0 into \textbf{Bi2Code}—a modular task execution interface with four key capabilities: (1) dual-arm API support, (2) decomposable and structured prompts, (3) standardized symbolic logging, and (4) embedded observation hooks for multimodal tracing. Notably, Bi2Code extends task coverage from 14 to 50, but we restrict evaluation to 10 overlapping tasks to ensure a fair comparison. The complete set of environment functions available in Bi2Code, along with a representative usage example, is provided in Appendix\ref{sec:appendix_available_functions} and~\ref{sec:appendix_example}, respectively.

We report three metrics to capture both one-shot accuracy and iterative repair efficiency:

\begin{itemize}
    \item \textbf{ASR} (Average Success Rate): Average task completion rate across all candidate executions.
    \item \textbf{Top5-ASR}: Success rate among the top-5 performing candidates.
    \item \textbf{CR-Iter}: Mean number of repair iterations to exceed 50\% success.
\end{itemize}

\begin{table*}[t]
\centering
\small
\begin{tabular}{lcccccc}
\toprule
\multirow{2}{*}{\textbf{Task}} & \multicolumn{3}{c}{\textbf{RoboTwin 1.0}} & \multicolumn{3}{c}{\textbf{Bi2Code}} \\
\cmidrule(lr){2-4} \cmidrule(lr){5-7}
 & Code as Policies & CodeAct & HyCodePolicy & Code as Policies & CodeAct & HyCodePolicy \\
\midrule
Beat Block Hammer        & 16\% & 48\% & \textbf{56\%} & 23\% & 34\% & 53\% \\
Handover Block            & 2\%  & 41\% & 45\% & 17\% & \textbf{50\%} & 27\% \\
Pick Diverse Bottles     & \textbf{65\%} & \textbf{65\%} & 64\% & 60\% & 60\% & 62\% \\
Pick Dual Bottles Easy  & 99\% & 99\% & \textbf{100\%} & \textbf{100\%} & \textbf{100\%} & \textbf{100\%} \\
Place Container Plate    & 66\% & 79\% & \textbf{91\%} & 84\% & 84\% & 82\% \\
Place Dual Shoes         & 19\% & 22\% & \textbf{25\%} & 0\%  & 2\%  & 22\% \\
Place Empty Cup          & 90\% & 90\% & \textbf{100\%} & 61\% & 61\% & 85\% \\
Place Shoe                & 72\% & 90\% & 90\% & \textbf{100\%} & \textbf{100\%} & \textbf{100\%} \\
Stack Blocks Three       & 1\%  & 2\%  & 4\%  & 76\% & 76\% & \textbf{82\%} \\
Stack Blocks Two         & 44\% & 68\% & 64\% & \textbf{100\%} & \textbf{100\%} & \textbf{100\%} \\
\bottomrule
\end{tabular}
\caption{\textbf{Task-Specific Performance Comparison of Different Feedback Mechanisms.} Average success rates for individual tasks are presented across `Code as Policies', `CodeAct', and `HyCodePolicy' variants, utilizing both RoboTwin 1.0 and Bi2Code interfaces. Bold numbers indicate the best result for each task.}
\label{tab:task_performance}
\vspace{2pt}
\end{table*}

They jointly evaluate program quality, repair effectiveness, and convergence efficiency across varying system architectures.

\subsection{Q1: How Efficient is Bi2Code Compared to Baseline RoboTwin 1.0?}

We first quantify the architectural impact of Bi2Code in a one-shot setting (\textit{i.e.}, Code as Policies). Table~\ref{tab:codegen_efficiency_transposed} shows that Bi2Code yields significantly shorter programs (569.4 vs.~1236.6 tokens), with reduced prompt length and higher structural similarity to human-written code. Crucially, it enables dual-arm parallelism via a unified API abstraction, which is absent in RoboTwin~1.0.

These improvements stem from the structured prompting and geometric API modularization designed into Bi2Code. Higher AST similarity (+21.06\%), CodeBERT similarity (+1.08\%), and Unixcoder alignment (+5.97\%) indicate that Bi2Code not only reduces code size but also improves semantic clarity and functional alignment. These properties are essential for subsequent feedback-based refinement, as modular and interpretable code facilitates localized repair. A detailed case study comparing HyCodePolicy-generated and human-written code under the Bi2Code interface is provided in Appendix~\ref{sec:appendix_case}.

\subsection{Q2: Do Feedback and Multimodal Repair Improve Performance? A Hierarchical Ablation Perspective}

\begin{figure}[t]
  \centering
  \includegraphics[width=\linewidth]{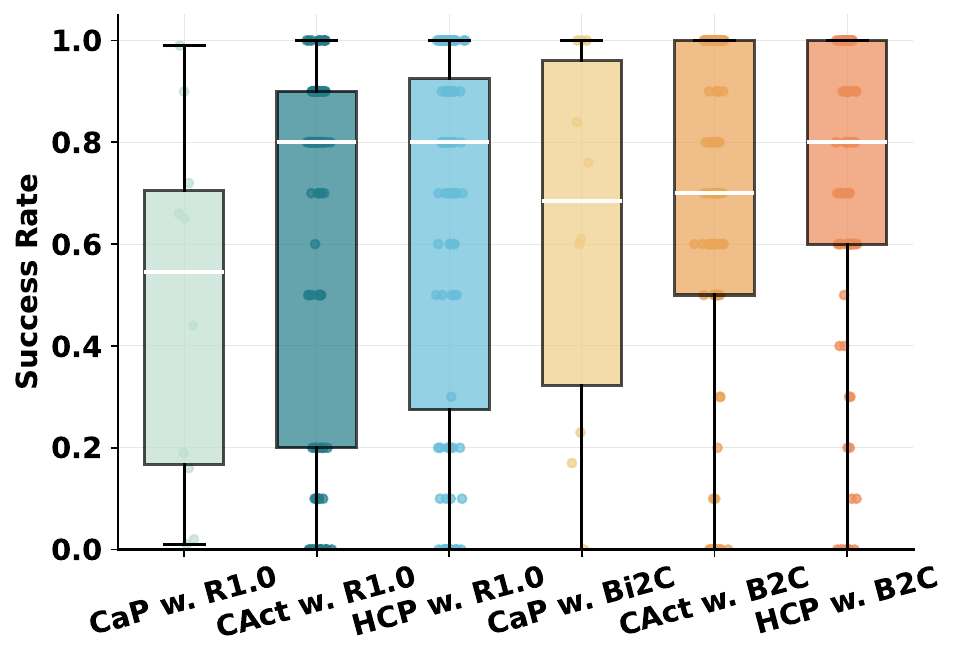}
  \vspace{-8pt}
  \caption{\textbf{Distribution of Task Success Rates.} The figure shows the distribution of success rates across all tasks for RoboTwin 1.0 and Bi2Code under different feedback configurations. `HyCodePolicy` in Bi2Code results in compact distributions centered above 80\%, with stronger worst-case performance.}
  \label{fig:robotwin-success-rate-distribution}
  \vspace{-2pt}
\end{figure}

To systematically assess the impact of feedback modalities, we adopt a hierarchical variant structure that lends itself naturally to ablation-style analysis. Each system variant—\textit{Code as Policies}, \textit{CodeAct}, and \textit{HyCodePolicy}—incrementally augments its predecessor with richer feedback capabilities, ranging from no feedback (one-shot execution) to symbolic correction and finally vision-language-grounded diagnosis. This layered design isolates the contribution of each feedback mechanism to both repair effectiveness and convergence speed. Detailed prompts and configurations for multimodal observation and iterative feedback are provided in Appendix~\ref{sec:appendix_Observation}.

\begin{figure*}[t]
  \centering
  \begin{minipage}{0.69\textwidth} 
    \centering
    \footnotesize
    \setlength{\tabcolsep}{2.2pt} 
    \begin{tabular}{@{}ll|ll|ll@{}}
    \toprule
    \textbf{Task} & \textbf{Rate} & \textbf{Task} & \textbf{Rate} & \textbf{Task} & \textbf{Rate} \\
    \midrule
    Adjust Bottle & 100\% & Beat Block Hammer & 53\% & Blocks Ranking Rgb & 80\% \\
    Blocks Ranking Size & 80\% & Click Alarmclock & 0\% & Click Bell & 10\% \\
    Dump Bin Bigbin & 0\% & Grab Roller & 74\% & Handover Block & 27\% \\
    Handover Mic & 0\% & Hanging Mug & 0\% & Lift Pot & 40\% \\
    Move Can Pot & 30\% & Move Pillbottle Pad & 50\% & Move Playingcard Away & 90\% \\
    Move Stapler Pad & 100\% & Open Laptop & 0\% & Open Microwave & 0\% \\
    Pick Diverse Bottles & 62\% & Pick Dual Bottles & 100\% & Place A2B Left & 50\% \\
    Place A2B Right & 60\% & Place Bread Basket & 0\% & Place Bread Skillet & 0\% \\
    Place Can Basket & 0\% & Place Cans Plasticbox & 100\% & Place Container Plate & 82\% \\
    Place Dual Shoes & 22\% & Place Empty Cup & 85\% & Place Fan & 70\% \\
    Place Burger Fries & 100\% & Place Mouse Pad & 100\% & Place Object Basket & 0\% \\
    Place Object Scale & 80\% & Place Object Stand & 90\% & Place Phone Stand & 0\% \\
    Place Shoe & 100\% & Press Stapler & 0\% & Put Bottles Dustbin & 0\% \\
    Put Object Cabinet & 0\% & Rotate Qrcode & 80\% & Scan Object & 0\% \\
    Shake Bottle & 0\% & Shake Bottle Horizontally & 0\% & Stack Blocks Three & 82\% \\
    Stack Blocks Two & 100\% & Stack Bowls Three & 20\% & Stack Bowls Two & 30\% \\
    Stamp Seal & 20\% & Turn Switch & 0\% & \textbf{Avg Success Rate} & \textbf{43.34\%} \\
    \bottomrule
    \end{tabular}
    \captionof{table}{\textbf{Per-Task Success Rates of HyCodePolicy on the Full Bi2Code Task Suite.} This table summarizes the average success rates for all 50 tasks supported by the Bi2Code interface when executed using our proposed HyCodePolicy framework.}
    \vspace{-2mm}
    \label{tab:success_rates_fourcol_compact}
  \end{minipage}
  \hfill
  \begin{minipage}{0.27\textwidth} 
    \centering
    \includegraphics[width=1\linewidth]{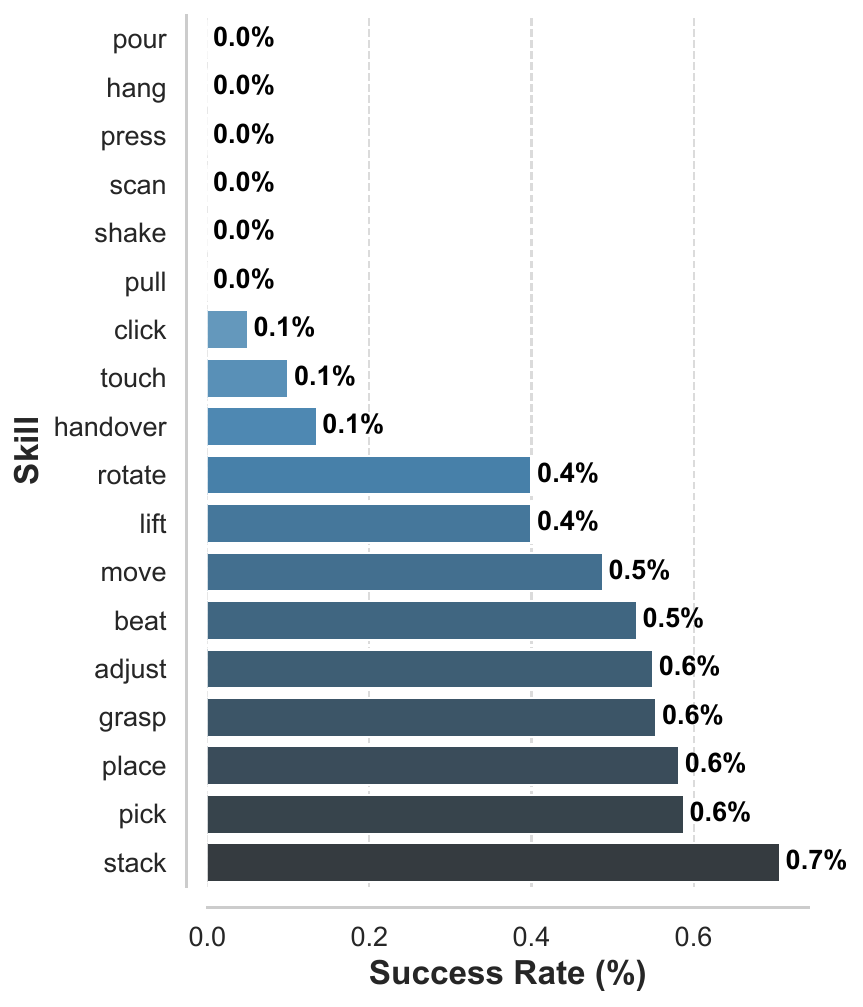}
    \caption{\textbf{Skill success rates}, where each skill's success rate is calculated as the average success rate of tasks that utilize the skill.}
    \label{fig:skill_success_rate}
  \end{minipage}
\end{figure*}

As shown in Figure~\ref{fig:robotwin-success-rate-distribution}, the task success rate distribution highlights the impact of different feedback mechanisms. In the Code as Policies setting for RoboTwin 1.0, there is high variance and a low median, indicating inconsistent performance. The introduction of CodeAct reduces this variance, improving overall consistency and shifting the central tendency upwards. The most notable improvement is observed with HyCodePolicy in Bi2Code, where success rates become more concentrated above 80\%.

Table~\ref{tab:robotwin_results} presents our main results. Moving from Code as Policies to CodeAct improves RoboTwin~1.0’s ASR from 47.4\% to 60.4\%, while HyCodePolicy further improves it to 63.9\%. On Bi2Code, the effect is even more pronounced, with ASR rising from 62.1\% to 66.7\% with CodeAct, and to 71.3\% with HyCodePolicy.

Importantly, HyCodePolicy reduces CR-Iter, indicating faster convergence. Bi2Code reaches functional success in 1.76 iterations, compared to 2.42 under RoboTwin~1.0. These gains highlight that HyCodePolicy not only provides richer feedback but also that Bi2Code’s structured code and logging better expose the failure loci for targeted correction.

From a system design perspective, this demonstrates a tight coupling between modular task APIs (Bi2Code), perceptual traceability (VLM hooks), and causally grounded repair. The observed improvements are not just empirical, but emerge directly from design choices that expose richer internal structure to the feedback loop.

\subsection{Q3: Which Part of the Planning Does Multimodal Feedback Matter Most?}

To dissect performance at a finer granularity, Table~\ref{tab:task_performance} compares per-task success rates under CodeAct and HyCodePolicy. Tasks like \textit{stack blocks three}, \textit{place empty cup}, and \textit{handover block} show large gains under HyCodePolicy. These tasks require accurate spatial reasoning, precise object alignment, and nuanced perception---factors poorly captured by symbolic logs.

By contrast, tasks such as \textit{pick dual bottles easy}, which rely on deterministic logic, show near-identical performance across all variants. This confirms that symbolic feedback suffices in low-ambiguity domains, while HyCodePolicy’s multimodal loop is crucial for disambiguating failure in visually complex settings.

These trends reinforce that HyCodePolicy’s strength lies in its perceptual introspection capability---leveraging visual observations not just for detection, but for actionable error attribution. This “why it failed” signal is critical to enabling effective revision in cases where symbolic traces are silent.

Performance differences between RoboTwin 1.0 and Bi2Code are not directly comparable in Table~\ref{tab:task_performance} , as they rely on distinct motion planning backends. While RoboTwin 1.0 uses a deterministic planner, Bi2Code adopts \texttt{Curobo}, whose planning process exhibits inherent stochasticity. This variability affects not only generated policies but also expert-authored programs, introducing confounding factors in direct cross-platform comparison.

\subsection{Q4: How Well Does HyCodePolicy Generalize Across Diverse Tasks?}

To evaluate the generalization capability of our proposed HyCodePolicy framework, we extend evaluation from the 10 core tasks (shared by RoboTwin1.0 and Bi2Code) to the full 50-task suite supported by Bi2Code. Crucially, the framework architecture, feedback logic, and prompting structure were jointly tuned only on the shared subset reported in Tab.~\ref{tab:task_performance}. No additional task-specific adaptation, hyperparameter change, or manual prompt adjustment was introduced when scaling to the remaining 40 tasks in Tab.~\ref{tab:success_rates_fourcol_compact}.

HyCodePolicy demonstrates strong zero-shot generalization, performing well in tasks involving structured placement, stacking, and planar manipulation (e.g., \textit{place mouse pad}, \textit{stack blocks two}, \textit{adjust bottle}), confirming that its framework abstractions, such as compositional prompts and perceptual repair, remain effective across diverse tasks.

However, notable failures occur in tasks requiring non-rigid object handling (\textit{place bread basket}), articulated motion (\textit{open microwave}), or temporal sequencing (\textit{press stapler}, \textit{scan object}). These shortcomings stem from limitations in the action API, world modeling, and trajectory-level reasoning, highlighting areas for future extension. Specifically, tasks like \textit{press}, \textit{scan}, \textit{shake}, \textit{pull}, and \textit{pour} have 0\% success due to the absence of these skills in the current API. While HyCodePolicy attempts to simulate these actions by adjusting parameters or combining actions, it often results in suboptimal performance. This occurs because human-designed policies typically involve specific poses, a capability HyCodePolicy currently lacks.

The skill success rate chart (Fig.~\ref{fig:skill_success_rate}) shows that tasks requiring complex or uncommon skills, such as \textit{press} (0\%) and \textit{scan} (0\%), perform poorly compared to more common skills like \textit{stack} (70.7\%) and \textit{pick} (58.8\%). This underscores HyCodePolicy's strength in basic manipulation tasks but reveals challenges in advanced, non-rigid manipulation and precise object control.

\vspace{5pt}
Overall, HyCodePolicy exhibits robust zero-shot generalization but its limitations suggest key areas for future improvement, particularly in expanding the action API to support more specialized skills for complex tasks.



\section{Conclusions and Limitations}\label{section:conclusions_limitations}

This paper introduced \textbf{HyCodePolicy}, a novel closed-loop architecture for language-grounded robotic manipulation. By integrating hierarchical subgoal decomposition, geometrically grounded program synthesis, multimodal monitoring, and iterative program repair, HyCodePolicy transforms code into a dynamic tool for perception, self-monitoring, and autonomous refinement. Our experimental evaluation using the \textbf{Bi2Code} interface demonstrated significant improvements in code conciseness, success rates, and convergence speed. The multimodal feedback, combining symbolic logs and VLM-based diagnostics, enhanced performance in tasks requiring spatial reasoning and perceptual disambiguation, marking a significant step toward more robust, interpretable, and autonomous robotic systems with reduced human supervision.

Despite these advancements, \textbf{HyCodePolicy} faces challenges in \textbf{articulated object manipulation}, \textbf{fine-grained parameter adjustments}, and \textbf{deformable object dynamics}. Tasks requiring precise arm poses and granular spatial understanding remain difficult, as do those involving complex temporal reasoning. Future work will focus on improving these areas, with an emphasis on enhancing temporal reasoning, integrating external knowledge, and developing lifelong learning mechanisms.

{
    \small
    \bibliographystyle{ieeenat_fullname}
    \bibliography{main}
}
\clearpage 
\appendix 

\section{Appendix} 

This appendix provides the detailed prompt structure and components used for guiding the LLM in generating robotic control code. The prompt is constructed from several distinct modules, each serving a specific purpose in defining the task, available tools, and contextual information. For detailed usage instructions, refer to \url{https://robotwin-platform.github.io/doc/usage/expert-code-gen.html}.

\subsection{LLM-Generated Code and Human-Written Code Case Study}
\label{sec:appendix_case}

\noindent\textbf{LLM-Generated Code (\texttt{gpt\_place\_shoe})}
\begin{lstlisting}[caption={LLM-Generated Code for \texttt{gpt\_place\_shoe} Task}]
class gpt_place_shoe(place_shoe):
    def play_once(self):
        # Initial observation
        self.save_camera_images(task_name="place_shoe", step_name="step1_initial_scene_state", generate_num_id="generate_num_0")
        
        # Get the shoe's position to determine which arm to use
        shoe_pose = self.shoe.get_pose()
        shoe_position = shoe_pose.p
        arm_tag = ArmTag("left" if shoe_position[0] < 0 else "right")
        
        # Grasp the shoe
        self.move(self.grasp_actor(actor=self.shoe, arm_tag=arm_tag, pre_grasp_dis=0.1, grasp_dis=0))
        self.save_camera_images(task_name="place_shoe", step_name="step2_shoe_grasped", generate_num_id="generate_num_0")
        
        # Lift the shoe up to avoid collision
        self.move(self.move_by_displacement(arm_tag=arm_tag, z=0.07, move_axis='world'))
        self.save_camera_images(task_name="place_shoe", step_name="step3_shoe_lifted", generate_num_id="generate_num_0")
        
        # Get target pose from the top functional point of the target block
        target_pose = self.target_block.get_functional_point(1, "pose")
        
        # Place the shoe on the target block with head towards left side
        self.move(self.place_actor(actor=self.shoe, arm_tag=arm_tag, target_pose=target_pose, functional_point_id=0, pre_dis=0.1, dis=0.02, is_open=True, constrain="align", pre_dis_axis='fp'))
        self.save_camera_images(task_name="place_shoe", step_name="step4_shoe_placed", generate_num_id="generate_num_0")
        
        # Lift the gripper up after placing
        self.move(self.move_by_displacement(arm_tag=arm_tag, z=0.07, move_axis='world'))
        
        # Return arm to origin
        self.move(self.back_to_origin(arm_tag=arm_tag))
        
        # Final observation
        self.move(self.save_camera_images(task_name="place_shoe", step_name="step5_final_scene_state", generate_num_id="generate_num_0"))
\end{lstlisting}

\noindent\textbf{Human-Written Code (\texttt{place\_shoe})}
\begin{lstlisting}[caption={Human-Written Code for \texttt{place\_shoe} Task}]
class place_shoe(base_task):
    def play_once(self):
        # Get the shoe's position to determine which arm to use
        shoe_pose = self.shoe.get_pose().p
        arm_tag = ArmTag("left" if shoe_pose[0] < 0 else "right")

        # Grasp the shoe with specified pre-grasp distance and gripper position
        self.move(self.grasp_actor(self.shoe, arm_tag=arm_tag, pre_grasp_dis=0.1, gripper_pos=0))

        # Lift the shoe up by 0.07 meters in z-direction
        self.move(self.move_by_displacement(arm_tag=arm_tag, z=0.07))

        # Get target_block's functional point as target pose
        target_pose = self.target_block.get_functional_point(0)

        # Place the shoe on the target_block with alignment constraint and specified pre-placement distance
        self.move(self.place_actor(self.shoe, arm_tag=arm_tag, target_pose=target_pose, functional_point_id=0, pre_dis=0.12, constrain="align"))
        
        # Open the gripper to release the shoe
        self.move(self.open_gripper(arm_tag=arm_tag))
\end{lstlisting}

The AI-generated code tends to be more verbose, explicitly logging intermediate visual states and detailing parameters (e.g., \texttt{pre\_dis\_axis='fp'}, \texttt{is\_open=True}), while human-written scripts are more minimal, omitting intermediate steps and favoring compact execution. Despite functional similarity, the structural differences illustrate that \textbf{MLLM-generated programs are not only executable but emphasize step-by-step clarity}, contributing to more robust feedback and repair.

\subsection{Prompt Templates, Code Template and Basic Info}
\label{sec:appendix_prompt}

\subsubsection{Overall Prompt Template}
The complete prompt is constructed by concatenating several key components, as shown in the template below. This structure ensures all necessary information—basic environment details, task description, available actors, API functions, and current code—is provided to the language model for generating or repairing code.

\begin{tcolorbox}[promptblock, title=\textbf{Prompt Template}]
\begin{verbatim}
Prompt = (
f"#Basic Info:\n{BASIC_INFO}\n"
f"#Task Description:\n{task_description}\n"
f"#Actor List:\n{actor_list}\n"
f"#Available API:\n{available_env_function}\n"
f"#Function Example:\n{function_example}\n"
f"#Current Code:\n{current_code}"
)
\end{verbatim}
\end{tcolorbox}

\subsubsection{\texttt{Basic Info}}
The \texttt{BASIC\_INFO} string provides fundamental details about the simulation environment, including units of measurement, pose representation, coordinate system conventions, and how to access functional points of actors. This foundational information is crucial for the language model to understand the operational context and correctly interpret geometric and interaction-related instructions.

\begin{tcolorbox}[promptblock, title=\textbf{BASIC\_INFO Constant}]
    \begin{minipage}{\linewidth} 
        \ttfamily\footnotesize 
        BASIC\_INFO = '''\\
        In this environment, distance 1 indicates 1 meter long. Pose is representated as 7 dimention, [x, y, z, qw, qx, qy, qz].\\
        For a 7-dimensional Pose object, you can use Pose.p to get the [x, y, z] coordinates and Pose.q to get the [qw, qx, qy, qz] quaternion orientation.\\
        All functions which has parameter actor, and all of actor should be in the Actor object.\\
        In the world coordinate system, the positive directions of the xyz coordinate axes are right, front, and upper respectively, so the direction vectors on the right, front,\\
        and upper sides are [1,0,0], [0,1,0], [0,0,1] respectively. In the same way, we can get the unit vectors of the left side, back side and down side.\\
        Each actor in the environment has one or more functional points, which are specific locations designed for interactions.\\
        Access functional points using actor.get\_functional\_point(point\_id, return\_type), where return\_type can be "pose", "p", or "q".\\
        '''
    \end{minipage}
\end{tcolorbox}

\subsubsection{\texttt{CODE\_TEMPLATE}}
The \texttt{CODE\_TEMPLATE} provides a basic Python class structure that the generated policies must adhere to. This template includes necessary imports and defines a `gpt\_$TASK_NAME$` class inheriting from `$TASK_NAME$`, ensuring the generated code integrates seamlessly into the existing simulation framework. The `play\_once` method is intended to be filled with the generated policy logic.

\begin{lstlisting}[caption=\textbf{CODE\_TEMPLATE Constant}, label={lst:code_template}]
CODE_TEMPLATE = '''
from envs._base_task import Base_Task
from envs.$TASK_NAME$ import $TASK_NAME$
from envs.utils import *
import sapien

class gpt_$TASK_NAME$($TASK_NAME$):
    def play_once(self):
        pass
'''
\end{lstlisting}

\subsection{Observation Agent Prompt}
\label{sec:appendix_Observation}

This section details the prompts used to guide the AI observation agent. The agent's role is twofold: first, to strategically insert observation (camera image capture) function calls into the robot task code at critical points; and second, to analyze the captured images to provide feedback on task execution. A third component outlines how multimodal observation feedback is incorporated into iterative code generation.

\subsubsection{Insert Observation Function Calls}
This prompt is designed to instruct the LLM to augment existing robot task code with calls to a camera observation function. The goal is to capture significant visual scene changes, providing a mechanism for step-by-step monitoring of the robot's execution.

\begin{tcolorbox}[promptblock, title=\textbf{Prompt for Inserting Observation Calls}]
\begin{minipage}{\linewidth}
\ttfamily\footnotesize
You are an expert in robot programming. I have a robot task code that needs observation functions added for monitoring.

Task information: \{task\_info\}

I need you to:

1. Identify ONLY the main logical steps in this task implementation that cause SIGNIFICANT SCENE CHANGES

2. After each such logical step in the code, insert a camera observation function with this format: \\
\hspace*{2em}self.save\_camera\_images(task\_name="\{task\_name\}", step\_name="stepX\_descriptive\_name", generate\_num\_id="generate\_num\_\{generate\_num\_id\}")

3. Provide a numbered list of all the steps you've identified in the task

4. ADD AN OBSERVATION AT THE BEGINNING OF THE TASK to capture the initial scene state

5. ADD AN OBSERVATION AT THE END OF THE TASK to capture the final scene state

Here's the current code: \textbf{python} \{task\_code\}

IMPORTANT CONSTRAINTS:

- ADD FEWER THAN 10 OBSERVATION POINTS in total

- ONLY add observations after operations that cause VISIBLE SCENE CHANGES

- Do NOT add observations for planning, calculations, or any operations that don't visibly change the scene

- Focus on key state changes like: robot arm movements, gripper operations, object manipulations

- Skip observations for intermediate movements, planning steps, or calculations

- The observation function is already defined in the code

- Give each step a descriptive name like "gripper\_closed", "move\_to\_target", etc.

- The step number (X in stepX) should increase sequentially

- DO NOT MODIFY ANY EXISTING ROBOT OPERATION CODE — only insert observation function calls after existing code without changing the original functionality

Format your response as follows:

STEP\_LIST:
1. First step description
2. Second step description
...

MODIFIED\_CODE:
python
<the entire modified code with observation functions inserted>
\end{minipage}
\end{tcolorbox}

\subsubsection{Observe Task Execution by Analyzing Step-by-Step Images}
This prompt guides the AI to analyze visual feedback (step-by-step images) from the robot's execution. The analysis focuses on identifying successful and failed steps, and providing detailed reasoning for any task failures.

\begin{tcolorbox}[promptblock, title=\textbf{Prompt for Image-Based Execution Analysis}]
\begin{minipage}{\linewidth}
\ttfamily\footnotesize
Analyze the execution of the following robot task: \\
Task name: \{task\_name\} \\
Task description: \{task\_info.get('description', 'No description provided')\} \\
Task goal: \{task\_info.get('goal', 'No goal provided')\}

You will be shown images from each step of the task execution. Please analyze:

1. Whether each step was executed successfully.

2. If any step failed, identify which one and explain why.

3. Whether the overall task was successfully completed.

4. If the task failed, provide detailed reasoning.

You will see execution images for the following steps: \{', '.join(step\_names)\}
\end{minipage}
\end{tcolorbox}

\subsubsection{Iterative Correction with Multimodal Observation Feedback}
When the initially generated code is unsuccessful, this prompt demonstrates how multimodal feedback (last error message and visual observation feedback) is incorporated to guide the iterative refinement of the robot task code.

\begin{tcolorbox}[promptblock, title=\textbf{Prompt for Iterative Correction with Feedback}]
\begin{minipage}{\linewidth}
\ttfamily\footnotesize
if observation\_feedback: \\
\hspace*{1em}prompt = ( \\
\hspace*{2em}f"The code is unsuccessful, \textbackslash n\# Last Error Message: \textbackslash n\{last\_error\}\textbackslash n\textbackslash n" \\
\hspace*{2em}f"\# Visual Observation Feedback: \textbackslash n\{observation\_feedback\}\textbackslash n\textbackslash n" \\
\hspace*{2em}f"\# Task Description: \textbackslash n\{task\_description\}\textbackslash n\textbackslash n" \\
\hspace*{2em}f"\# Actor List: \textbackslash n\{actor\_list\}\textbackslash n\textbackslash n" \\
\hspace*{2em}\# Additional components like available API, function example, and current code would follow \\
\hspace*{2em}\# f"\#Available API:\textbackslash n\{available\_env\_function\}\textbackslash n" \\
\hspace*{2em}\# f"\#Function Example:\textbackslash n\{function\_example\}\textbackslash n" \\
\hspace*{2em}\# f"\#Current Code:\textbackslash n\{current\_code\}" \\
\hspace*{1em})
\end{minipage}
\end{tcolorbox}

\subsection{Available Environment Functions}
\label{sec:appendix_available_functions}

The following \texttt{AVAILABLE\_ENV\_FUNCTION}, details the various robotic control functions exposed to the language model. Each entry provides the function's signature, a brief description of its purpose, and a comprehensive list of its parameters. These functions form the foundational API for generating and executing robot manipulation policies within the simulation environment.

\begin{tcolorbox}[promptblock, title=\textbf{Available API}]
\small 
The following functions are available for the AI model to use for controlling the robotic arms and interacting with the environment:
\begin{itemize}[leftmargin=*, noitemsep, topsep=0pt, partopsep=0pt] 
    \item \textbf{\texttt{open\_gripper(self, arm\_tag: ArmTag, pos=1.) -> tuple[ArmTag, list[Action]]}} \\
    Opens the gripper of the specified arm. Returns: tuple[ArmTag, list[Action]] containing the gripper-open action. Args: arm\_tag: Which arm's gripper to open; pos: Gripper position (1 = fully open).
    \item \textbf{\texttt{close\_gripper(self, arm\_tag: ArmTag, pos=0.) -> tuple[ArmTag, list[Action]]}} \\
    Closes the gripper of the specified arm. Returns: tuple[ArmTag, list[Action]] containing the gripper-close action. Args: arm\_tag: Which arm's gripper to close; pos: Gripper position (0 = fully closed).
    \item \textbf{\texttt{move(self, actions\_by\_arm1: tuple[ArmTag, list[Action]], actions\_by\_arm2: tuple[ArmTag, list[Action]] = None)}} \\
    Executes action sequences on one or both robotic arms simultaneously. No Return. Args: actions\_by\_arm1: Action sequence for the first arm, formatted as (arm\_tag, [action1, action2, ...]); actions\_by\_arm2: Optional, action sequence for the second arm.
    \item \textbf{\texttt{move\_by\_displacement(self, arm\_tag: ArmTag, z=0., move\_axis='world') -> tuple[ArmTag, list[Action]]}} \\
    Moves the end-effector of the specified arm along relative directions and sets its orientation. Returns: tuple[ArmTag, list[Action]] containing the move-by-displacement actions. Args: arm\_tag: The arm to control; z: Displacement along the z-axis (in meters); move\_axis: 'world' means displacement is in world coordinates, 'arm' means displacement is in local coordinates.
    \item \textbf{\texttt{grasp\_actor(self, actor: Actor, arm\_tag: ArmTag, pre\_grasp\_dis=0.1, grasp\_dis=0, gripper\_pos=0., contact\_point\_id=None) -> tuple[ArmTag, list[Action]]}} \\
    Generates a sequence of actions to pick up the specified Actor. Returns: tuple[ArmTag, list[Action]] containing the grasp actions. Args: actor: The object to grasp; arm\_tag: Which arm to use; pre\_grasp\_dis: Pre-grasp distance (default 0.1 meters), the arm will move to this position first; grasp\_dis: Grasping distance (default 0 meters), the arm moves from the pre-grasp position to this position and then closes the gripper; gripper\_pos: Gripper closing position (default 0, fully closed); contact\_point\_id: Optional list of contact point IDs; if not provided, the best grasping point is selected automatically.
    \item \textbf{\texttt{place\_actor(self, actor: Actor, arm\_tag: ArmTag, target\_pose: list | np.ndarray, functional\_point\_id: int = None, pre\_dis=0.1, dis=0.02, is\_open=True, **kwargs) -> tuple[ArmTag, list[Action]]}} \\
    Places a currently held object at a specified target pose. Returns: tuple[ArmTag, list[Action]] containing the place actions. Args: actor: The currently held object; arm\_tag: The arm holding the object; target\_pose: Target position/orientation, It is recommended to use the return value of actor.get\_functional\_point(..., 'pose') or pose in actor\_list as target\_pose; functional\_point\_id: Optional ID of the functional point; if provided, aligns this point to the target, otherwise aligns the base of the object; pre\_dis: Pre-place distance (default 0.1 meters), arm moves to this position first; dis: Final placement distance (default 0.02 meters), arm moves from pre-place to this location, then opens the gripper; is\_open: Whether to open the gripper after placing (default True), Set False if you need to keep gripper closed to maintain hold of the object; **kwargs: Other optional parameters: constrain : {'free', 'align', 'auto'}, default='auto' Alignment strategy: 'free': Only forces the object's z-axis to align with the target point's z-axis, other axes are determined by projection. 'align': Forces all axes of the object to align with all axes of the target point. 'auto': Automatically selects a suitable placement pose based on grasp direction (vertical or horizontal). pre\_dis\_axis : {'grasp', 'fp'} or np.ndarray or list, default='grasp'. Specifies the pre-placement offset direction.
    \item \textbf{\texttt{back\_to\_origin(self, arm\_tag: ArmTag) -> tuple[ArmTag, list[Action]]}} \\
    Returns the specified arm to its predefined initial position. Returns: tuple[ArmTag, list[Action]] containing the return-to-origin action. Args: arm\_tag: The arm to return to origin.
\end{itemize}
\end{tcolorbox}

\subsection{Function Example}
\label{sec:appendix_example}

\noindent\texttt{FUNCTION\_EXAMPLE} provides practical examples and guidelines for using the available API functions. It demonstrates how to interact with actors, control grippers, and execute complex manipulation sequences, including single-arm and dual-arm operations. This section is essential for understanding the practical application of the API in generating robot control policies.

\begin{tcolorbox}[promptblock, title=\textbf{Function Examples}]
\small 

You can directly use the actors provided in the \texttt{actor\_list}:
\begin{lstlisting}
# For example, if actor_list contains ["self.object1", "self.object2"]
# You can directly use:
object1 = self.hammer
object2 = self.block
\end{lstlisting}

\noindent Using \texttt{ArmTag} class to represent arms:
\begin{lstlisting}
arm_tag = ArmTag("left")  # Left arm
arm_tag = ArmTag("right") # Right arm
\end{lstlisting}

\noindent Example of selecting an arm based on conditions:
\begin{lstlisting}
arm_tag = ArmTag("left" if actor_position[0] < 0 else "right")
\end{lstlisting}

\noindent Each actor in the environment may have multiple functional points that are useful for different interactions. Functional points provide precise locations for interactions like grasping, placing, or aligning objects.

\noindent To get a functional point from an actor:
\begin{lstlisting}
functional_point_pose = actor.get_functional_point(point_id, "pose")  % Returns a complete 7-dimensional Pose object
position = functional_point_pose.p  % [x, y, z] position
orientation = functional_point_pose.q % [qw, qx, qy, qz] quaternion orientation
\end{lstlisting}
Note: The pose from a functional point is already aligned for the task. Do NOT manually construct or rotate a quaternion.

\noindent When stacking one object on top of another (e.g., placing \texttt{blockA} on top of \texttt{blockB}):
\begin{lstlisting}
target_pose = self.last_actor.get_functional_point(point_id, "pose")
self.move(
    self.place_actor(
        actor=self.current_actor,
        target_pose=target_pose,
        arm_tag=arm_tag,
        functional_point_id=0,
        pre_dis=0.1,
        dis=0.02,
        pre_dis_axis="fp"
    )
)
\end{lstlisting}

\noindent For actors already of type \texttt{Pose}, such as \texttt{actor\_pose}, you do \textbf{not} need to call \texttt{.get\_pose()} again:
\begin{lstlisting}
self.move(
    self.place_actor(
        self.box,
        target_pose=self.actor_pose,
        arm_tag=grasp_arm_tag,
        functional_point_id=0,
        pre_dis=0,
        dis=0,
        is_open=False,
        constrain="free",
        pre_dis_axis='fp',
    )
)
\end{lstlisting}
Note: For \texttt{target\_actor}, you must use \texttt{get\_pose()} or \texttt{get\_functional\_point()}.

\noindent To select an arm and grasp based on actor position:
\begin{lstlisting}
actor_pose = self.actor.get_pose()
actor_position = actor_pose.p

arm_tag = ArmTag("left" if actor_position[0] < 0 else "right")

self.move(
    self.grasp_actor(actor=self.actor, arm_tag=arm_tag)
)
\end{lstlisting}

\noindent Example grasping and lifting:
\begin{lstlisting}
self.move(
    self.grasp_actor(
        actor=self.actor,
        arm_tag=arm_tag,
        pre_grasp_dis=0.1,
        grasp_dis=0
    )
)

self.move(
    self.move_by_displacement(
        arm_tag=arm_tag,
        z=0.07,
        move_axis='world'
    )
)
\end{lstlisting}

\noindent Gripper control:
\begin{lstlisting}
self.move(self.open_gripper(arm_tag=arm_tag, pos=1.0))  # Fully open
self.move(self.open_gripper(arm_tag=arm_tag, pos=0.5))  # Half open
self.move(self.close_gripper(arm_tag=arm_tag, pos=0.0)) # Fully close
self.move(self.close_gripper(arm_tag=arm_tag, pos=0.5)) # Half close
\end{lstlisting}

\noindent Example placing:
\begin{lstlisting}
self.move(
    self.place_actor(
        actor=self.actor,
        arm_tag=arm_tag,
        target_pose=self.target_pose,
        functional_point_id=0,
        pre_dis=0.1,
        dis=0.02,
        is_open=True,
        pre_dis_axis='fp',
    )
)

self.move(
    self.move_by_displacement(
        arm_tag=arm_tag,
        z=0.07,
        move_axis='world'
    )
)
\end{lstlisting}

\noindent Aligning a functional point with a target:
\begin{lstlisting}
self.move(
    self.place_actor(
        actor=self.actor,
        arm_tag=arm_tag,
        target_pose=target_pose,
        functional_point_id=0,
        pre_dis=0.1,
        dis=0.02,
        pre_dis_axis='fp'
    )
)
\end{lstlisting}

\noindent Move both arms simultaneously:
\begin{lstlisting}
left_arm_tag = ArmTag("left")
right_arm_tag = ArmTag("right")
self.move(
    self.grasp_actor(actor=self.left_actor, arm_tag=left_arm_tag),
    self.grasp_actor(actor=self.right_actor, arm_tag=right_arm_tag)
)

self.move(
    self.move_by_displacement(arm_tag=left_arm_tag, z=0.07),
    self.move_by_displacement(arm_tag=right_arm_tag, z=0.07)
)
\end{lstlisting}

\noindent Place left object while returning right arm to origin:
\begin{lstlisting}
move_arm_tag = ArmTag("left")
back_arm_tag = ArmTag("right")
self.move(
    self.place_actor(
        actor=self.left_actor,
        arm_tag=move_arm_tag,
        target_pose=target_pose,
        pre_dis_axis="fp"
    ),
    self.back_to_origin(arm_tag=back_arm_tag)
)
\end{lstlisting}

\noindent Return arms to initial positions:
\begin{lstlisting}
self.move(self.back_to_origin(arm_tag=arm_tag))

left_arm_tag = ArmTag("left")
right_arm_tag = ArmTag("right")
self.move(
    self.back_to_origin(arm_tag=left_arm_tag),
    self.back_to_origin(arm_tag=right_arm_tag)
)
\end{lstlisting}

\end{tcolorbox}


\end{document}